\documentclass[10pt,twocolumn,letterpaper]{article}

\usepackage{iccv}              

\usepackage{algorithm}
\usepackage{algorithmic} 
\usepackage{amsfonts}
\usepackage{multirow}
\usepackage{bm}
\usepackage{booktabs}

\definecolor{iccvblue}{rgb}{0.21,0.49,0.74}
\usepackage[pagebackref,breaklinks,colorlinks,allcolors=iccvblue]{hyperref}

\def\paperID{7119} 
\def\confName{ICCV}
\def\confYear{2025}

\title{DIVE: Taming DINO for Subject-Driven Video Editing}

\author{Yi Huang$^{1, 2}$ \quad Wei Xiong$^{3\ddagger}$ \quad He Zhang$^{4}$ \quad Chaoqi Chen$^{5}$
\\ 
Jianzhuang Liu$^{1}$ \quad Mingfu Yan$^{6, 1}$ \quad Shifeng Chen$^{1, 7\dagger}$
\\ 
$^1$Shenzhen Institutes of Advanced Technology, Chinese Academy of Sciences,\\ $^2$vivo AI Lab,
$^3$NVIDIA, $^4$Adobe Research, $^5$Shenzhen University, \\$^6$Southeast University, $^7$Shenzhen University of Advanced Technology\\
{\tt\small \{yi.huang, mf.yan, jz.liu, shifeng.chen\}@siat.ac.cn } \\
{\tt\small wxiongur@gmail.com, he.zhang92@rutgers.edu, cqchen1994@gmail.com} \\
}

\newcommand{\teaser}{
\vspace{-30pt}
\begin{center}
	\includegraphics[width=0.98\textwidth]{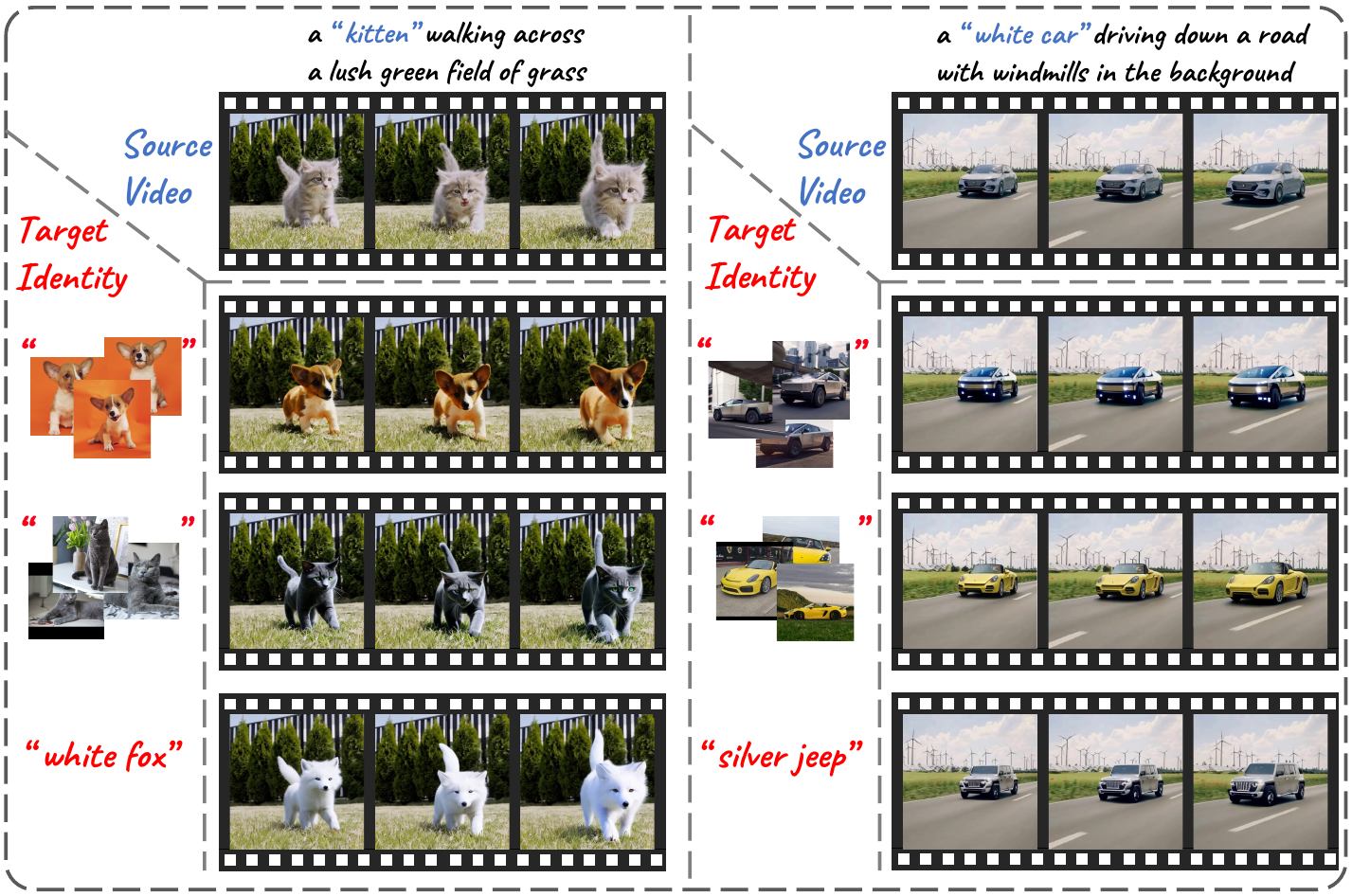}
\end{center}
\vspace{-15pt}
\captionof{figure}{Subject-driven video editing results achieved by DIVE. Given a source video, DIVE can accurately capture the subject’s motion trajectory and perform precise editing with various target identities, specified either by reference images or text prompts.}
\vspace{12pt}
\label{fig:teaser}
}

\begin{document}

\twocolumn[{%
	\vspace{-2em}
	\maketitle%
	
	\teaser%
}]

\begingroup
\renewcommand{\thefootnote}{}
\footnotetext{$^{\ddagger}$ Project lead. \quad \quad $^{\dagger}$ Corresponding author.}
\endgroup

\begin{abstract}
Building on the success of diffusion models in image generation and editing, video editing has recently gained substantial attention. However, maintaining temporal consistency and motion alignment still remains challenging. To address these issues, this paper proposes DINO-guided Video Editing (DIVE), a framework designed to facilitate subject-driven editing in source videos conditioned on either target text prompts or reference images with specific identities. The core of DIVE lies in leveraging the powerful semantic features extracted from a pretrained DINOv2 model as implicit correspondences to guide the editing process. Specifically, to ensure temporal motion consistency, DIVE employs DINO features to align with the motion trajectory of the source video. For precise subject editing, DIVE incorporates the DINO features of reference images into a pretrained text-to-image model to learn Low-Rank Adaptations (LoRAs), effectively registering the target subject’s identity. Extensive experiments on diverse real-world videos demonstrate that our framework can achieve high-quality editing results with robust motion consistency, highlighting the potential of DINO to contribute to video editing. 
\end{abstract} 
\vspace{-10pt}
\section{Introduction}
\label{sec:intro}

In recent years, diffusion-based generative models have emerged as a leading approach in the domain of Artificial Intelligence Generated Content (AIGC) due to their outstanding capabilities in capturing data distribution and their inherent resistance to mode collapse~\cite{sohl2015deep, ho2020denoising, song2021denoising, song2019generative, song2021score}. Extensive research has shown the exceptional potential of diffusion models in generating diverse and high-quality images conditioned on text prompts, a process commonly referred to as text-to-image (T2I) synthesis~\cite{dhariwal2021diffusion, ho2022cascaded, saharia2022photorealistic, rombach2022high}. Building upon this success, researchers have extended the application of T2I diffusion models to the field of text-to-video (T2V) generation~\cite{ho2022video, singer2023make, lv2024gpt4motion, ho2022imagen, ge2023preserve, xing2023survey, wei2024dreamvideo} and editing, aiming to achieve similar advancements in video content creation.

Compared to image editing~\cite{huang2024diffusion, brooks2023instructpix2pix, sheynin2023emu, li2024magiceraser}, video editing~\cite{sun2024diffusion, yang2023rerender, zhao2023controlvideo, qi2023fatezero, liu2024video, ouyang2024i2vedit} presents unique challenges, especially in capturing the motion from source videos and transferring it to edited results while maintaining temporal consistency. Most methods address this challenge by relying on pretrained T2I models, introducing additional temporal layers, motion modules, or unique mechanisms such as spatial-causal attention and temporal attention~\cite{wu2023tune, liu2024video, ku2024anyv2v, geyer2023tokenflow}. These methods typically transfer the source video's motion to the target video by injecting temporal attention maps or features. For instance, Tune-A-Video~\cite{wu2023tune} performs cross frame attention between each frame and anchor frames (e.g., the first and previous frames) to preserve appearance consistency.  TokenFlow~\cite{geyer2023tokenflow} further explicitly propagates diffusion features based on inter-frame correspondences to improve consistency. However, the stored features in these extra modules or internal blocks may inadvertently retain some source appearance details, leading to a blend of source and target appearances and resulting in visual incoherence.

To better disentangle motion from appearance in source videos, some methods utilize video correspondences that carry minimal appearance information, such as flow~\cite{liang2024flowvid, cong2023flatten}, depth~\cite{kara2024rave, esser2023structure, liew2023magicedit}, or edge~\cite{yang2023rerender, zhao2023controlvideo} map sequences, for temporal motion modeling. For instance, RAVE~\cite{kara2024rave} utilizes depth maps extracted from source videos, which are then applied to generate target videos through ControlNet~\cite{zhang2023adding}. However, due to the high density of these sequences, existing techniques for extracting or estimating them often result in visual content incoherence and flickering. VideoSwap~\cite{gu2024videoswap} addresses this by observing that a subject's motion trajectory can be effectively represented using a small number of semantic points and utilizes these as sparse video correspondences to align the motion trajectory. Although this approach can accurately align motion trajectories, it requires accurate manual definition of these points in the source videos, which is labor-intensive.

This paper aims to establish more robust video correspondences that not only contain minimal appearance information but also can accurately align motion trajectories without manual annotation. Motivated by recent advances in self-supervised learning, we leverage a pretrained DINOv2 model~\cite{oquab2023dinov2}, a vision Transformer distilled using a large collection of natural images. DINO features have been validated to be robust and discriminative in capturing fine-grained semantic information, proving valuable across visual tasks~\cite{amir2021deep, melas2022deep, shtedritski2023learning}. Additionally, Zhang \textit{et al.}~\cite{zhang2023tale} reveal that DINO features excel at obtaining sparse and accurate matches on image semantic correspondence. However, their application in the video domain has not been fully explored. Based on these insights, we introduce DINO to the video domain and find that DINO features across video frames exhibit high semantic similarity, indicating their potential as robust video correspondences to guide the editing process.

To this end, we propose \textbf{DI}NO-guided \textbf{V}ideo \textbf{E}diting (DIVE), a framework that enables subject editing in source videos conditioned on either text prompts or reference images with specific identities, as shown in Figure~\ref{fig:teaser}. DIVE consists of three stages: temporal motion modeling, subject identity registration, and inference. In the first stage, we extract the DINO features from the source video and align them into the diffusion feature space through learnable MLPs to serve as motion guidance. In the second stage, we incorporate the DINO features of reference images into a pretrained T2I model to learn Low-Rank Adaptations (LoRAs) for identity guidance. In the final inference stage, we use DDIM inversion to obtain the latent noise of the source video and replace the source subject with the target subject in the text prompt, utilizing both the motion and identity guidance learned from the previous two stages.
Our contributions can be summarized as follows:
\begin{itemize}
	\item We show that tamed DINO features offer strong and robust correspondences for capturing temporal motion.
	\item We propose DIVE, a subject-driven video editing framework that leverages DINO features for consistent motion modeling and precise subject identity registration.
	\item Extensive experiments demonstrate that DIVE can achieve high-quality editing results while maintaining temporal consistency and motion alignment.
\end{itemize}

\section{Related Work}
\label{sec:related}

\subsection{Diffusion-Based Video Generation and Editing}
Diffusion models have recently emerged in various vision tasks, especially in image generation and editing~\cite{rombach2022high, zhang2023adding, mou2023t2i, brooks2023instructpix2pix, huang2023smartedit}. Building on their success, video generation and editing with diffusion models have attracted significant research interest. Early efforts focus primarily on training video diffusion models directly in pixel or latent spaces~\cite{ho2022video, ho2022imagen, blattmann2023align, wang2023lavie}, which require large datasets~\cite{xue2022advancing, wang2023internvid}. However, recent studies have shifted towards training-free models to reduce computational costs. For instance, Text2Video-Zero~\cite{khachatryan2023text2video} leverages the pretrained Stable Diffusion model for video synthesis, employing cross-attention with the first frame to maintain frame consistency.

As focus shifts from video generation to editing, research has increasingly explored methods that adapt pretrained T2I models, driven by their strong generative capabilities~\cite{wu2023tune, geyer2023tokenflow, ku2024anyv2v, li2024video, wang2024cove, zhang2024towards, li2024vidtome, cohen2024slicedit}. These methods often introduce extra temporal layers, motion modules, or spatial-temporal attention to ensure temporal coherence. For example, Tune-A-Video~\cite{wu2023tune} adds temporal self-attention and updates projection matrices in each attention block to enhance temporal modeling. FateZero~\cite{qi2023fatezero} stores attention maps at each inversion stage and fuses them during editing to preserve motion and structure. Others rely on auxiliary sequences, such as flow~\cite{liang2024flowvid, cong2023flatten, hu2023videocontrolnet}, depth~\cite{feng2024ccedit, kara2024rave, esser2023structure, liew2023magicedit}, or edge~\cite{yang2023rerender, zhao2023controlvideo} maps, to model motion. FLATTEN~\cite{cong2023flatten}, for instance, uses flow-guided attention to align patches across frames. However, these dense signals often cause flickering and incoherence. VideoSwap~\cite{gu2024videoswap} mitigates this with sparse semantic points as correspondences, but it requires manual point definition. In contrast, we leverage self-supervised DINOv2 features as implicit correspondences, enabling accurate motion alignment without manual annotation.

\subsection{Subject-Driven Image Generation}
Subject-driven image generation aims to create personalized images that consistently preserve the identity and unique characteristics of a given subject, typically guided by a few reference images. Earlier efforts in this domain, such as Textual Inversion~\cite{gal2023image} and DreamBooth~\cite{ruiz2023dreambooth}, lay the foundation for personalized generation~\cite{voronov2023loss, wei2023elite, liu2023cones, chen2023subject, shi2023instantbooth, tewel2023key, chen2023disenbooth, gu2023photoswap, yuan2023customnet, lu2023specialist, ye2023ip}. Textual Inversion~\cite{gal2023image} achieves this by learning a unique identifier word that represents a new subject, integrating it into the text encoder’s dictionary to maintain subject consistency. DreamBooth~\cite{ruiz2023dreambooth}, on the other hand, fine-tunes the entire Imagen~\cite{saharia2022photorealistic} model with a few reference images, associating a new, rare word with the specific subject to ensure the generated images retain the subject’s identity. To improve the representation of multiple new concepts simultaneously, Custom Diffusion~\cite{kumari2023multi} optimizes cross-attention parameters within Stable Diffusion, enabling joint training to combine multiple concepts without compromising subject identity. COMCAT~\cite{xiao2023comcat} further improves efficiency by introducing a low-rank approximation of attention matrices, significantly reducing storage requirements while maintaining high fidelity in the generated outputs. In addition, methods such as adapters, LoRAs, and their variants~\cite{gu2023mix, ruiz2024hyperdreambooth, kong2024omg} have become increasingly popular in subject-driven generation, offering parameter-efficient fine-tuning that allows for precise control over the personalized output. These advancements enable refined control over the generation process, enhancing the precision and accuracy with which the subject’s identity is maintained across generated images. In this paper, we aim to register the subject’s identity for editing source videos, either conditioned on text prompts or reference images.

\section{Method}
In this section, we begin by presenting the motivation behind our use of DINO feature correspondences in Section~\ref{sec:motivation}. Next, we provide an overview of the DIVE pipeline for subject-driven editing in Section~\ref{sec:dive_pipeline}, where we explain the technical details of each stage.

\begin{figure}[t]
	\small
	\centering
	\includegraphics[width=\linewidth]{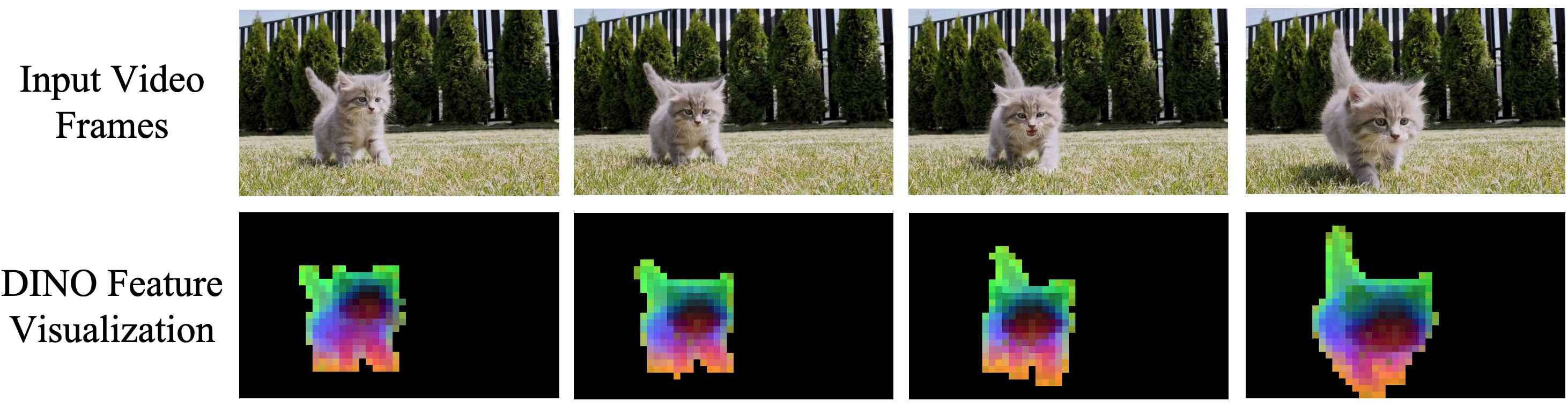}
	\vspace{-15pt}
	\caption{Visualization of DINO features across video frames, with background removed for clearer observation.}
	\label{dino_vis}
	\vspace{-10pt}
\end{figure}

\begin{figure*}[t]
	\small
	\centering
	\includegraphics[width=\textwidth]{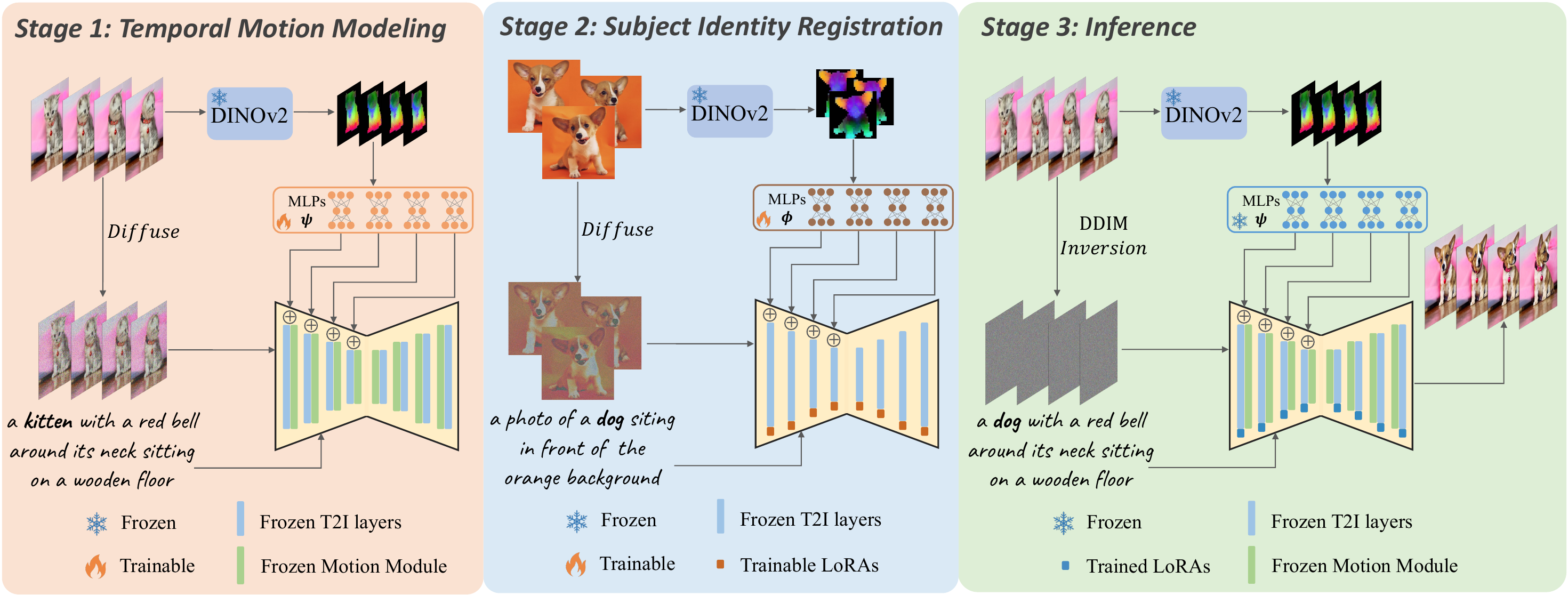}
	\vspace{-15pt}
	\caption{Overall pipeline of DIVE. It consists of three primary stages: temporal motion modeling, subject identity registration, and inference. In the first stage, we extract the DINO features from the source video and align them into the diffusion feature space through learnable MLPs to serve as motion guidance. In the second stage, we incorporate the DINO features of reference images into a pretrained T2I model to learn a Low-Rank Adaptation (LoRA) for identity guidance. During the final inference stage, we use DDIM inversion to obtain the latent noise of the source video and replace the source subject with the target subject in the text prompt, using both the motion and identity guidance learned from the previous two stages.}
	\label{pipeline}
	\vspace{-10pt}
\end{figure*}

\subsection{Motivation}
\label{sec:motivation}

A major challenge in subject-driven video editing is maintaining the motion trajectory of the source subject across frames. Existing methods often rely on auxiliary correspondences like optical flow or depth maps~\cite{liang2024flowvid, cong2023flatten, kara2024rave, esser2023structure}, but they tend to result in visual flickering due to the high density of these correspondences and can cause mismatches in the subject’s parts due to their lack of fine-grained semantics. To address these limitations, we explore DINO features as a more robust form of video correspondence, inspired by recent studies in image representation learning~\cite{zhang2023tale, tang2023emergent}.

To evaluate the potential of DINO features in video editing, we visualize them across consecutive frames using Principal Component Analysis (PCA). As shown in Figure~\ref{dino_vis}, the DINO features of a moving subject (a kitten) reveal three key advantages. First, they effectively track the subject’s motion across frames, consistently following the kitten’s trajectory. Second, they contain minimal appearance information, abstracting away fine details like texture and color to focus on the subject’s overall structure. Finally, they exhibit strong semantic consistency across frames while remaining discriminative within each frame. For example, the general shape and position of the DINO features stay consistent as the kitten moves, while distinct activations represent different parts of the subject (e.g., head, body, legs). This allows for precise, part-level matching in subject-driven video editing.

\subsection{DIVE}
\label{sec:dive_pipeline}
The overall pipeline of DIVE is presented in Figure~\ref{pipeline}, designed as a framework for subject-driven video editing that ensures temporal consistency and precise identity preservation. It consists of three primary stages: temporal motion modeling, subject identity registration, and inference, which are detailed in the following sections.

\subsubsection{Stage 1: Temporal Motion Modeling}
As illustrated in Figure~\ref{pipeline}, given a source video $\bm{V}$ containing $N$ frames, we first encode each frame into successive latent variables $\bm{Z} = \{\bm{z}_1, \bm{z}_2, \cdots, \bm{z}_N\}$ using a VAE encoder, where $\bm{Z} \in \mathbb{R} ^{N \times H\times W\times d}$, $H$ and $W$ represent the height and width of each latent variable, and $d$ denotes the dimension of each token. To simulate the diffusion process, we add random Gaussian noise at timestep $t$ to each latent frame, resulting in a noisy sequence $\bm{Z}^t = \{\bm{z}_1^t, \bm{z}_2^t, \cdots, \bm{z}_N^t\}$, which is then processed by the pretrained T2I model $\bm{\epsilon}_\theta$ (e.g., Stable Diffusion). Additionally, we incorporate motion layers from AnimateDiff~\cite{guo2023animatediff} into $\bm{\epsilon}_\theta$ to maintain essential temporal consistency across frames. In the U-Net encoder of the inflated T2I model $\bm{\epsilon}_{\theta'}^{enc}$, we extract intermediate features after each downsample block, represented as:
\begin{equation}
	\bm{F}^t = \{\bm{F}_1^t , \bm{F}_2^t , \bm{F}_3^t, \bm{F}_4^t\} = {\bm{\epsilon}_{\theta'}^{enc}(\bm{Z}^t)}.
\end{equation}
Consequently, the size of each feature $\bm{F}_l^t$ at downsample stage $l \in \{1, 2, 3, 4\}$ is $N\times H/2^l \times W/2^l \times d_l$, where $d_l$ denotes the channel dimension.

To capture the motion of the source video’s subject, we use the DINOv2 model to extract semantic features from each frame of $\bm{V}$. To avoid interference from background elements, we isolate the foreground subject features $\bm{F}_{d}\in \mathbb{R} ^{N \times h\times w\times c}$ by adaptively generating masks $\bm{M}$ from the complete features. Specifically, we apply PCA to reduce the dimensionality of the full DINO features to 1, and use thresholding to separate the foreground from the background. Due to the high semantic similarity and appearance sparsity across frames, $\bm{F}_{d}$ serves as effective temporal motion guidance. To integrate $\bm{F}_{d}$ back into the diffusion space, we introduce a set of learnable MLPs $\bm{\psi}=\{\bm{\psi}_l|l\in\{1,2,3,4\}\}$, following a similar technique used in VideoSwap~\cite{gu2024videoswap}. Each $\bm{\psi}_l$ projects $\bm{F}_d$ to match the feature dimension of the corresponding $\bm{F}_l^t$, followed by a resizing operation to align the spatial size to $H/2^l \times W/2^l $, as represented as:
\begin{equation}
	\bm{F}_l^s = Resize(\bm{\psi}_l(\bm{F}_d)), l \in \{1, 2, 3, 4\}.
	\label{eq:projection}
\end{equation}
Finally, $\bm{F}_l^s$ is added element-wise to the corresponding intermediate feature $\bm{F}_l^t$ of the U-Net encoder:
\begin{equation}
	\bm{F}_l^t \gets \bm{F}_l^t + \lambda\bm{F}_l^s, l \in \{1, 2, 3, 4\},
	\label{eq: injection_stage1}
\end{equation}
where $\lambda$ is a weighting parameter controlling the guidance strength. Further technical details on the projection and fusion process are provided in the supplementary materials.

The objective is to optimize the MLPs $\bm{\psi}$ to improve motion guidance from the DINO features, formulated as:
\begin{equation}
	\footnotesize
	\min_{\bm{\psi}} \bm{E}_{ \bm{\epsilon}\sim \bm{N}(\bm{0}, \bm{I}), t\sim U(T_{min}, T)}||[\bm{\epsilon}-\bm{\epsilon}_{\theta'}(\bm{Z}^t, t, \bm{c}, \bm{\psi}(\bm{F}_d))]\odot \bm{M}||^2_2,
	\label{eq:objective_stage1}
\end{equation}
where $\bm{c}$ represents the text prompt embedding, and $\bm{M}$ denotes binary masks that activate only the subject area. In this optimization process, $\lambda$ is set to 1, and $T_{min}$ is set to $T/2$ to emphasize learning at higher timesteps, reducing overfitting to the subject’s low-level details.

\subsubsection{Stage 2: Subject Identity Registration}

Beyond modeling the temporal motion of the source subject in stage 1, it is crucial to capture the target subject’s identity for precise subject replacement. In this stage, we use several (usually 3 to 5) reference images of the target subject. Similar to stage 1, we first transform these reference images into the latent space, adding Gaussian noise at timestep $t$ to produce noisy image latents $\bm{I}^t=\{\bm{I}_1^t, ..., \bm{I}_P^t\}$, where $P$ is the number of images. These noisy latents are then passed through the pretrained T2I model $\bm{\epsilon}_\theta$, from which we extract intermediate features $\bm{\tilde{F}}^t = \{\bm{\tilde{F}}_1^t, \bm{\tilde{F}}_2^t, \bm{\tilde{F}}_3^t, \bm{\tilde{F}}_4^t\} = {\bm{\epsilon}_\theta^{enc}(\bm{I}^t)}$ after each downsampling block of the U-Net encoder.

Since previous studies~\cite{tang2023emergent, zhang2023tale} suggest that Stable Diffusion features primarily capture low-level spatial information, whereas DINO features excel at extracting high-level semantic information and can achieve sparse yet accurate matches. Therefore, we also extract semantic features of the reference images using the DINOv2 model and fuse them with the internal diffusion model features to provide accurate identity guidance in the subsequent inference stage. Specifically, we mask out background elements to focus on the foreground subject as well. The resulting foreground semantic features, $\bm{\tilde{F}_d}$, are integrated back into the diffusion space through a set of learnable MLPs $\bm{\phi}$, yielding projected features $\bm{\tilde{F}}_l^s$, similar to Eq.~\ref{eq:projection}. We then fuse $\bm{\tilde{F}}_l^s$ with $\bm{I}^t$ via element-wise addition, as in Eq.~\ref{eq: injection_stage1}.
To effectively register the target subject’s identity, we employ Low-Rank Adaptations (LoRAs), denoted as $\Delta\theta$, rather than fine-tuning the entire T2I model. The optimization objective for identity registration is thus formulated as:
\begin{equation}
	\footnotesize
	\min_{\bm{\phi}, \Delta{\theta}} \bm{E}_{ \bm{\epsilon}\sim \bm{N}(\bm{0}, \bm{I}), t\sim U(1, T)}||[\bm{\epsilon}-\bm{\epsilon}_{\theta+\Delta{\theta}}(\bm{I}^t, t, \bm{\tilde{c}}, \bm{\phi}(\bm{\tilde{F}}_d))]\odot \bm{\tilde{M}}||^2_2,
	\label{eq:objective_stage2}
\end{equation}
where $\bm{\tilde{c}}$ represents the text prompt embedding, and $\bm{\tilde{M}}$ is a binary mask activating only the target subject’s region. Unlike in stage 1, this optimization is applied across all timesteps from $t=1$ to $t=T$. In this way, we register the target subject’s identity into the LoRAs by combining high-level semantic features from the DINOv2 model with low-level details from the pretrained T2I model, forming a robust representation of the target subject’s characteristics.

\subsubsection{Stage 3: Inference}

In the final stage, we synthesize the edited video frames by leveraging the motion and identity guidance learned in the previous stages. First, we perform DDIM inversion on each frame of the source video to obtain the initial latent noise representations, providing a starting point for the editing process. To guide the edited frames with the source video’s motion, we apply the motion features \(\bm{F}_l^s\) learned in stage 1, which are derived from the DINO-extracted semantic features of the source video. These features are added element-wise to the intermediate layers of the U-Net encoder during denoising the steps from $T$ to $T/2$. The MLPs $\bm{\psi}$, already optimized in stage 1 to project DINO-based motion guidance into the diffusion model space, are directly applied in inference mode here, preserving the motion trajectory of the original video. 
For identity guidance, we use the pretrained LoRAs \(\Delta \theta\) from stage 2 to integrate the target subject’s identity into the model without further feature fusion or additional MLP projections. These learned LoRAs encode the characteristics of the target subject directly within the T2I model, ensuring that the target identity is accurately and consistently represented across frames. To preserve the unedited background in the source video, we apply latent blending using a foreground mask, generated from P2P~\cite{hertz2022prompt}, to merge the latent variables from inversion and denoising at each timestep. Additionally, we replace the source subject word in the text prompt (e.g., “cat”) with the target subject identity word (e.g., “dog”) to ensure alignment with the target identity. 

As the denoising process progresses, the combined motion and identity guidance together work on each frame to produce an edited video that retains the original motion and accurately replaces the subject with the target identity from the reference images.
\section{Experiments}
\label{sec:experiments}

\begin{figure*}[t]
	\small
	\centering
	\includegraphics[width=\textwidth]{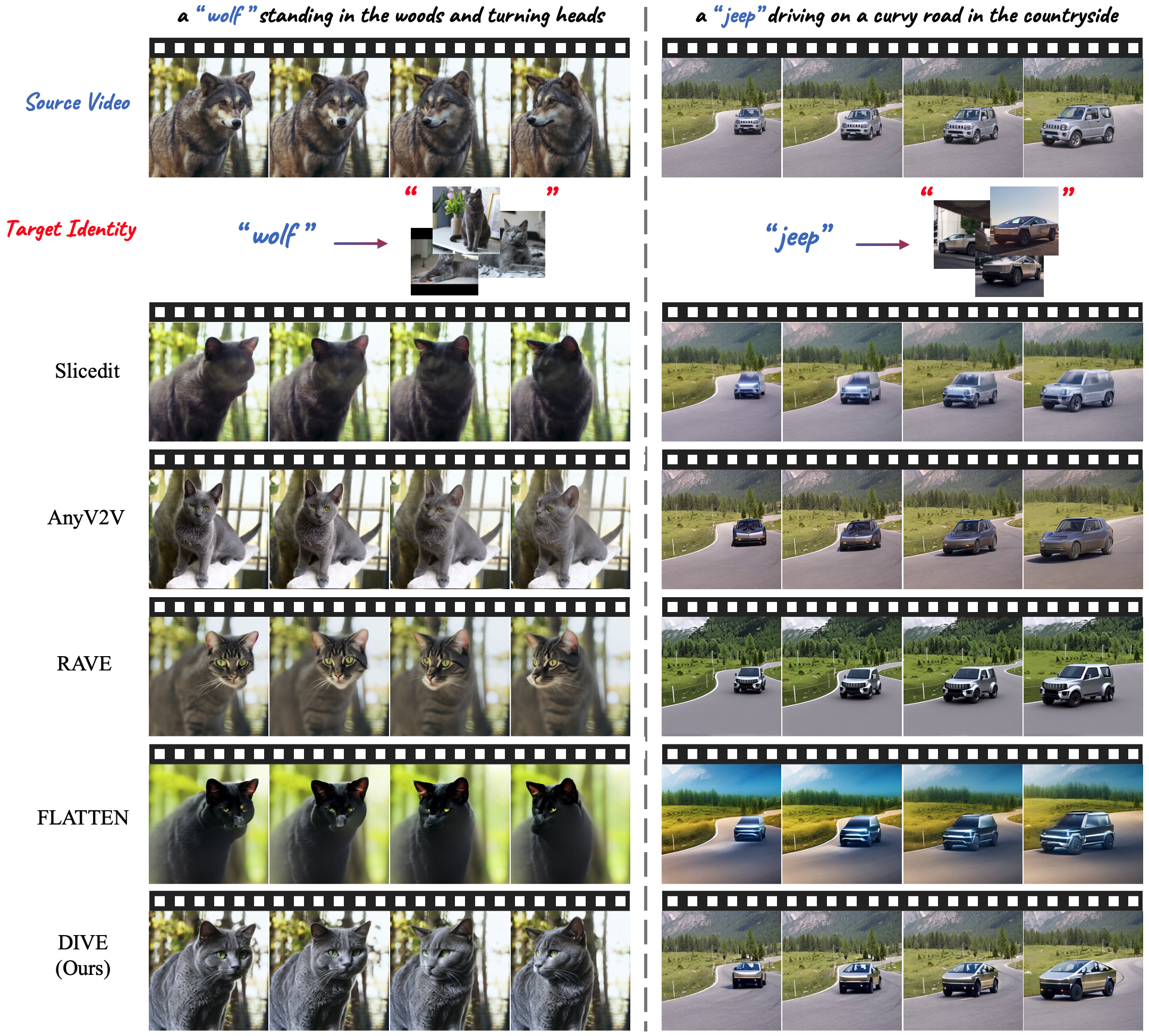}
	\vspace{-15pt}
	\caption{Qualitative comparison of DIVE with various editing methods, where target identities are specified by reference images. DIVE can accurately replace the source subject with target identity while maintaining both motion trajectory and background context.}
	\label{ref_comp}
	\vspace{-10pt}
\end{figure*}

\subsection{Experimental Setup}
For the model architecture, we adopt Stable Diffusion 1.5 from the official Huggingface repository, incorporating the pretrained motion layer from AnimateDiff~\cite{guo2023animatediff} as the foundational model. For the DINOv2 model, we utilize the ViT-g/14 variant without registers\footnote{https://github.com/facebookresearch/dinov2}.

We evaluate our method on a dataset of 30 videos curated from DAVIS~\cite{perazzi2016benchmark} and the Internet, covering both objects and animals. Each video sample consists of 16 frames with a temporal stride of 4 to match the temporal window of the motion layer in AnimateDiff. We crop and reisze these videos to two alternate resolutions for editing ($H \times W$): \(512 \times 512\) and \(512 \times 896\). For videos at \(512 \times 512\), we further resize them to \(448 \times 448\) for DINO feature extraction. This preprocessing allows us to match feature scales accurately. For instance, the DINO features have a spatial dimension of \(16 \times 16\), while the encoder features \(\bm{F}^t = \{\bm{F}_1^t , \bm{F}_2^t , \bm{F}_3^t, \bm{F}_4^t\}\) of the Stable Diffusion U-Net are at scales of \(\{64 \times 64, 32 \times 32, 16 \times 16, 8 \times 8\}\). Similarly, for \(512 \times 896\) videos, we resize them to a \(448 \times 768\) for DINO feature extraction correspondingly. 
We further collect 8 target identity categories (4 for animals and 4 for objects), with each category containing 3 to 5 reference images curated from DreamBench~\cite{ruiz2023dreambooth} and the Internet. The reference images undergo the same preprocessing steps as the videos. Beyond image guidance, we also perform text-guided subject editing on the source videos for a comprehensive comparison with other methods.

In terms of implementation settings, in stage 1, we use the Adam optimizer with a learning rate of \(5 \times 10^{-4}\), optimizing for 50 to 100 iterations. In stage 2, we use the Adam optimizer with a learning rate of \(1 \times 10^{-4}\) and optimize for 800 to 1000 iterations. In stage 3, we employ 50 steps of DDIM inversion and 50  steps of denoising, with $\lambda$ in Eq.~\ref{eq: injection_stage1} set to 1.0. All experiments are conducted on one single NVIDIA RTX 3090 GPU.

\begin{figure*}[t]
	\small
	\centering
	\includegraphics[width=\textwidth]{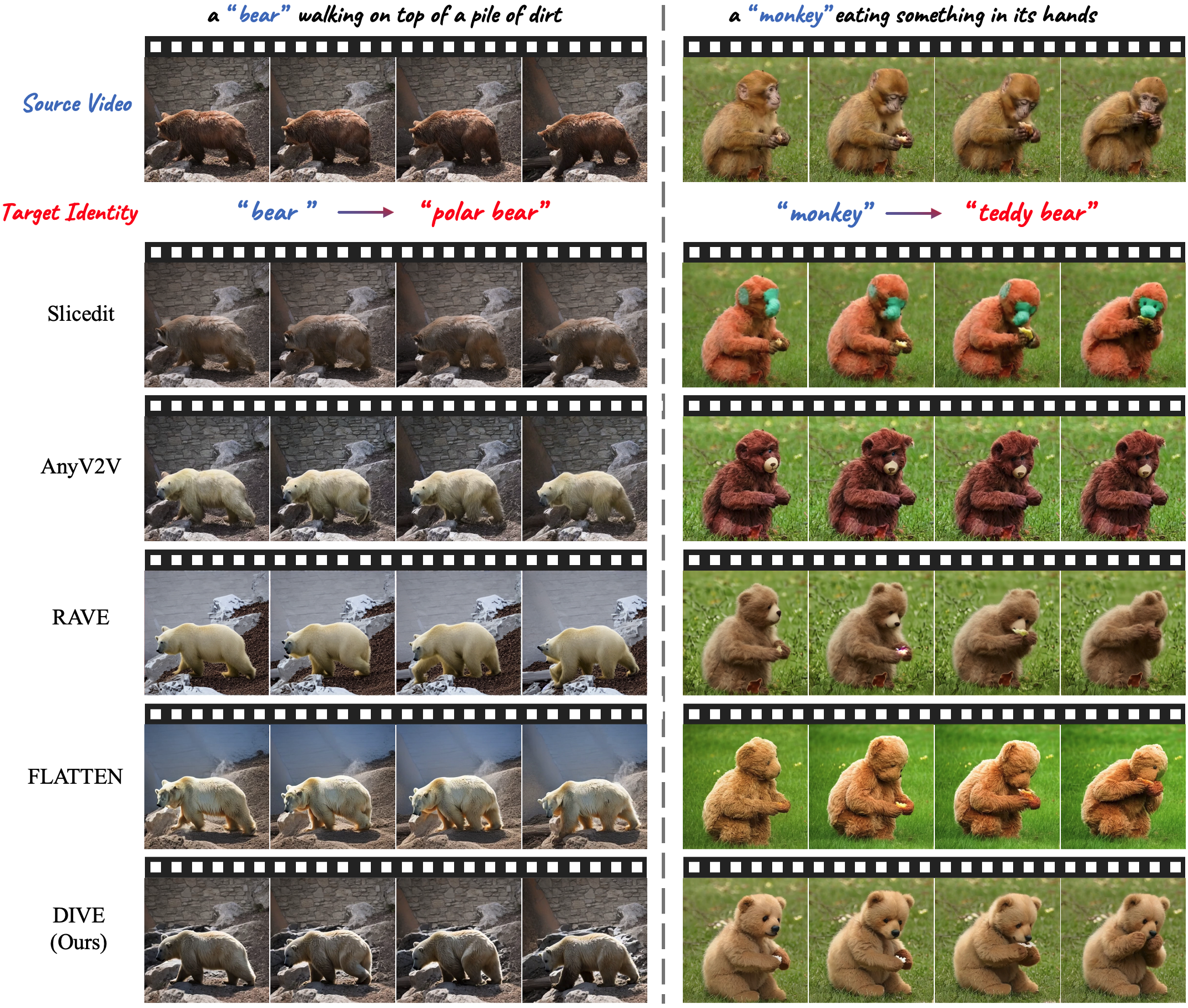}
	\vspace{-15pt}
	\caption{Qualitative comparison of DIVE with various editing methods, where target identities are specified by text prompts. DIVE outperforms previous methods with superior visual quality and temporal consistency.}
	\label{text_comp}
	\vspace{-10pt}
\end{figure*}

\subsection{Qualitative Comparison}
We compare DIVE against 4 baseline methods: Slicedit~\cite{cohen2024slicedit}, AnyV2V~\cite{ku2024anyv2v}, FLATTEN~\cite{cong2023flatten}, and RAVE~\cite{kara2024rave}, where the former two are attention-based and the latter two are correspondence-based. 
The comparison results on various videos with different reference images are shown in Figure~\ref{ref_comp}. As observed, DIVE can accurately replaces the source subject’s identity with that of the target based on reference images while aligning with the source motion trajectory and preserving the background. In contrast, Slicedit and AnyV2V struggle to maintain motion alignment and often introduce appearance details from the source subject, as simple attention transfer techniques are limited in disentangling motion and appearance effectively. Although RAVE and FLATTEN model the source motion using correspondences like optical flows and depth maps, they tend to produce blurred and inconsistent results due to the high density of these correspondences.

Additionally, to demonstrate DIVE’s generalization to text-guided video editing, we omit the identity registration step (stage 2 in Figure~\ref{pipeline}) and directly replace the source subject with the target one in the text prompt during inference in stage 3. Visual comparisons with other methods are presented in Figure~\ref{text_comp}, where DIVE shows superior temporal consistency by aligning the source motion across all parts of the subject and achieving coherent content across frames. More results are provided in the supplementary materials.

\subsection{Quantitative Comparison}
For quantitative evaluation, we use several metrics to assess DIVE and the baseline methods, which are (1) Text Alignment~\cite{wu2023cvpr}, by computing the text-image similarity in the CLIP feature space for each frame-prompt pair; (2) Image Alignment~\cite{kumari2023multi}, by measuring the visual similarity between each edited frame and the reference images using CLIP image features; (3) Temporal Consistency~\cite{wu2023cvpr}, by calculating the similarity of CLIP features across consecutive frames; and (4) Overall Video Quality~\cite{huang2024vbench}, by calculating scores from multiple dimensions and averaging them. Additionally, we conduct a user study with 50 participants, asking them to select their preferred results among these methods. 
The results listed in Table~\ref{tab_comp}, demonstrate that DIVE consistently outperforms other methods across all quantitative metrics and achieves high preference rates in the user study. This superior performance highlights DIVE’s ability to produce high-quality, temporally consistent subject edits that align well with the given text prompts or reference images.

\begin{table}[!tb]
	\centering
	\resizebox{\linewidth}{!}{
		\begin{tabular}{c|ccccc}
			\toprule
			\multirow{2}{*}{Method}
			& \multicolumn{1}{c}{\multirow{2}{*}{\begin{tabular}[c]{@{}c@{}}Text\\ Alignment~\cite{wu2023cvpr} $\uparrow$\end{tabular}}} 
			& \multicolumn{1}{c}{\multirow{2}{*}{\begin{tabular}[c]{@{}c@{}}Image\\ Alignment~\cite{kumari2023multi} $\uparrow$\end{tabular}}}
			& \multicolumn{1}{c}{\multirow{2}{*}{\begin{tabular}[c]{@{}c@{}}Temporal\\ Consistency~\cite{wu2023cvpr} $\uparrow$\end{tabular}}} 
			& \multicolumn{1}{c}{\multirow{2}{*}{\begin{tabular}[c]{@{}c@{}}Overall\\ Video Quality~\cite{huang2024vbench} $\uparrow$\end{tabular}}} 
			& \multicolumn{1}{c}{\multirow{2}{*}{\begin{tabular}[c]{@{}c@{}}User \\ Study $\uparrow$\end{tabular}}} \\ 
			~& \multicolumn{1}{c}{} & \multicolumn{1}{c}{} & \multicolumn{1}{c}{} & \multicolumn{1}{c}{} & \multicolumn{1}{c}{}\\
			\toprule
			~&\multicolumn{5}{c}{\textbf{Reference Image Guided Subject Editing}}\\
			\midrule
			Slicedit~\cite{cohen2024slicedit} &28.21 & 64.57 & 91.09& 0.592 & 6.73$\%$ \\
			AnyV2V~\cite{ku2024anyv2v} & 28.13 & 78.26 & 90.52 &0.439 & 13.2$\%$ \\
			FLATTEN~\cite{cong2023flatten} & 28.79 & 69.32 & 92.09& 0.683& 8.67$\%$ \\
			RAVE~\cite{kara2024rave} & 28.26 & 66.25 & 91.71 & 0.646 & 5.80$\%$\\
			\textbf{DIVE (Ours)} & \textbf{29.43} & \textbf{84.27} & \textbf{92.33}  & \textbf{0.775} & \textbf{65.6$\textbf{\%}$}\\
			\toprule
			~&\multicolumn{5}{c}{\textbf{Text Guided Subject Editing}}\\
			\midrule
			Slicedit~\cite{cohen2024slicedit} & 31.24 & $\backslash$ & 92.95 &0.562  & 5.50$\%$  \\ 
			AnyV2V~\cite{ku2024anyv2v} & 31.05 & $\backslash$ & 93.73 &0.533 & 19.9$\%$  \\ 
			FLATTEN~\cite{cong2023flatten} & 31.55 & $\backslash$ & 95.35 &0.567& 11.9$\%$  \\
			RAVE ~\cite{kara2024rave}& 31.57 & $\backslash$ & 95.12 &0.588& 9.90$\%$  \\
			\textbf{DIVE (Ours)} & \textbf{32.29} & $\backslash$& \textbf{95.89} &\textbf{0.614}&  \textbf{52.8$\textbf{\%}$ } \\ 
			\bottomrule
	\end{tabular}
}
	\vspace{-5pt}
	\caption{Quantitative evaluation for DIVE and other methods on three objective metrics and user study, under both reference image and text-guided video subject editing.}
	\vspace{-10pt}
	\label{tab_comp}
\end{table}

\subsection{Ablation Study}
\begin{figure}[t]
	\small
	\centering
	\includegraphics[width=\linewidth]{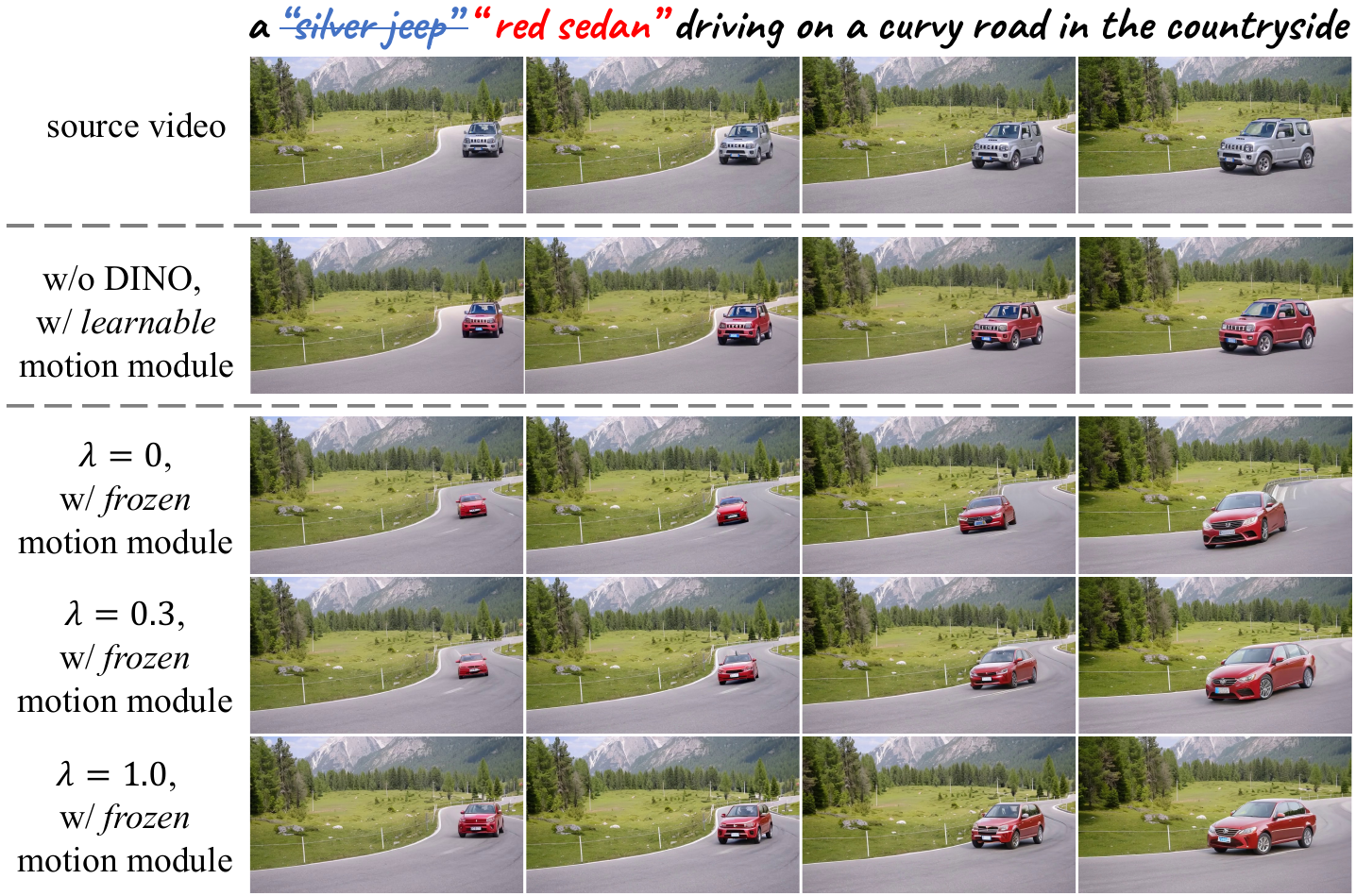}
	\vspace{-15pt}
	\caption{Visual ablations of motion guidance in stage 1.}
	\label{abla_temp}
	\vspace{-5pt}
\end{figure}

\noindent \textbf{Motion Guidance.}
To validate the effectiveness of DINO features as temporal motion guidance in stage 1, we remove the DINO feature extraction branch and the learnable MLPs \(\bm{\psi}\) in Eq.~\ref{eq: injection_stage1}, but fine-tune the motion module from AnimateDiff~\cite{guo2023animatediff}. The editing result shown in the second row of Figure~\ref{abla_temp} and Table~\ref{tab_abla} indicates that although this setup can capture the source motion, it struggles to disentangle the motion from the appearance of the source subject, leading to image misalignment. In contrast, our method can fully edit the subject without introducing elements of the source appearance, as DINO features exhibit high sparsity and semantic consistency across frames (see the last row). Additionally, to further evaluate the effectiveness of DINO features in capturing source motion trajectories, we vary the guidance injection weighting parameter \(\lambda\) in Eq.~\ref{eq: injection_stage1} while keeping the motion module frozen. The comparison in  Table~\ref{tab_abla} and the last three rows of Figure~\ref{abla_temp} show that without adequate DINO feature guidance, the motion cannot be consistently maintained, demonstrating the robustness of DINO features as video correspondences for reliable video editing.

\begin{figure}[t]
	\small
	\centering
	\includegraphics[width=\linewidth]{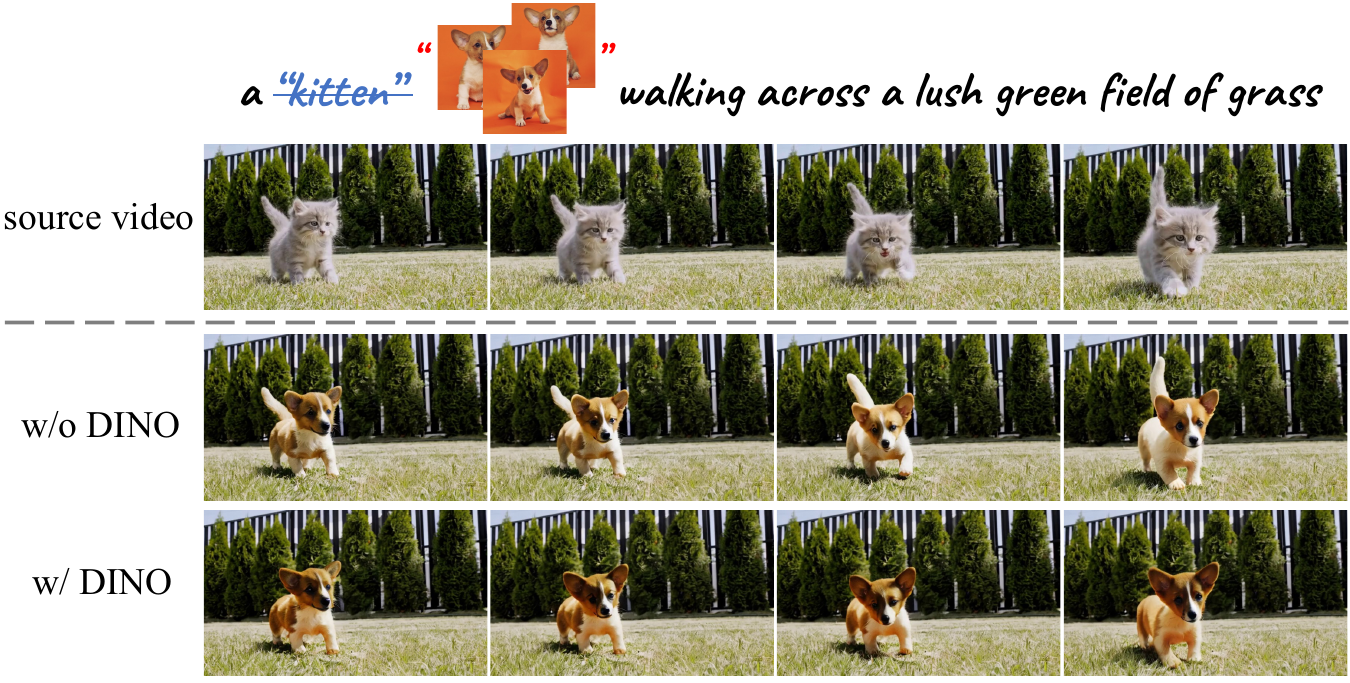}
	\vspace{-15pt}
	\caption{Visual ablations of identity guidance in stage 2.}
	\label{abla_id}
	\vspace{-3pt}
\end{figure}

\noindent \textbf{Identity Guidance.}
To evaluate the impact of DINO features on identity registration in stage 2, we perform an ablation study with and without DINO guidance. Results are shown in Figure~\ref{abla_id} and Table~\ref{tab_abla}. Without DINO guidance, subjects in edited videos tend to be less aligned with reference images. For example, the edited corgi incorrectly has a tail, though the reference images show a tail-less corgi. This error occurs because, without DINO’s semantic guidance, the model relies on biases from the pretrained diffusion model. In contrast, with DINO guidance, the edited corgi correctly matches the reference images. This demonstrates that DINO features provide accurate, part-level guidance, allowing the model to follow reference images closely and achieve high semantic fidelity to the target identity.

\begin{table}[t]
	\centering
	\resizebox{\linewidth}{!}{
		\begin{tabular}{c|cccc}
			\toprule
			\multirow{2}{*}{Implementation}
			& \multicolumn{1}{c}{\multirow{2}{*}{\begin{tabular}[c]{@{}c@{}}Text\\ Alignment~\cite{wu2023cvpr} $\uparrow$\end{tabular}}} 
			& \multicolumn{1}{c}{\multirow{2}{*}{\begin{tabular}[c]{@{}c@{}}Image\\ Alignment~\cite{kumari2023multi} $\uparrow$\end{tabular}}}
			& \multicolumn{1}{c}{\multirow{2}{*}{\begin{tabular}[c]{@{}c@{}}Temporal\\ Consistency~\cite{wu2023cvpr} $\uparrow$\end{tabular}}} 
			& \multicolumn{1}{c}{\multirow{2}{*}{\begin{tabular}[c]{@{}c@{}}Overall\\ Video Quality~\cite{huang2024vbench} $\uparrow$\end{tabular}}}  \\ 
			~& \multicolumn{1}{c}{} & \multicolumn{1}{c}{} & \multicolumn{1}{c}{} & \multicolumn{1}{c}{} \\
			\toprule
			~&\multicolumn{4}{c}{\textbf{Motion Guidance}}\\
			\midrule
			w/o DINO, & \multirow{2}*{29.91} &\multirow{2}*{67.49} & \multirow{2}*{92.18}  & \multirow{2}*{0.737} \\
w/ \textit{learnable} motion module &~&~&~&~  \\
			\midrule
			$\lambda=0$, & \multirow{2}*{28.78} &\multirow{2}*{79.50} & \multirow{2}*{87.24}  & \multirow{2}*{0.581} \\
w/  \textit{frozen} motion module &~&~&~&~  \\
\midrule
			$\lambda=0.3$, & \multirow{2}*{29.05} &\multirow{2}*{80.28} & \multirow{2}*{89.53}  & \multirow{2}*{0.628} \\
w/  \textit{frozen} motion module &~&~&~&~  \\
\midrule
			$\lambda=1.0$, & \multirow{2}*{\textbf{29.43}} &\multirow{2}*{\textbf{84.27}} & \multirow{2}*{\textbf{92.33}}  & \multirow{2}*{\textbf{0.775}} \\
w/  \textit{frozen} motion module &~&~&~&~  \\
			\toprule
			~&\multicolumn{4}{c}{\textbf{Identity Guidance}}\\
			\midrule
			w/o DINO & 29.25 & 77.36 & 92.12 &0.722   \\ 
			\midrule
			w/ DINO & \textbf{29.43} & \textbf{84.27} & \textbf{92.33}  & \textbf{0.775}   \\ 
			\bottomrule
		\end{tabular}
	}
	\vspace{-5pt}
	\caption{Quantitative ablations of motion and identity guidance.}
	\vspace{-10pt}
	\label{tab_abla}
\end{table}
\section{Conclusion}
\label{sec:conclusion}
This paper presents DIVE, a framework for subject-driven video editing that leverages DINO features as robust video correspondences to achieve reliable motion alignment and precise identity registration. DIVE effectively replaces the subject in a source video with a specified target identity, whether described by reference images or text prompts, while preserving the original motion trajectory and background context. Extensive qualitative and quantitative comparisons demonstrate that DIVE outperforms state-of-the-art methods in visual quality, motion consistency, and identity alignment, highlighting the potential of DINO to contribute to video editing.


\section*{Acknowledgement}
\label{sec:acknowledgement}
This work was supported by the Shenzhen Science and Technology Program (JSGG20220831105002004) and Shenzhen Key Laboratory of Computer Vision and Pattern Recognition.

{
	\small
	\bibliographystyle{ieeenat_fullname}
	\bibliography{main}

\begin{thebibliography}{79}
\providecommand{\natexlab}[1]{#1}
\providecommand{\url}[1]{\texttt{#1}}
\expandafter\ifx\csname urlstyle\endcsname\relax
  \providecommand{\doi}[1]{doi: #1}\else
  \providecommand{\doi}{doi: \begingroup \urlstyle{rm}\Url}\fi

\bibitem[Amir et~al.(2021)Amir, Gandelsman, Bagon, and Dekel]{amir2021deep}
Shir Amir, Yossi Gandelsman, Shai Bagon, and Tali Dekel.
\newblock Deep vit features as dense visual descriptors.
\newblock \emph{arXiv preprint arXiv:2112.05814}, 2\penalty0 (3):\penalty0 4,
  2021.

\bibitem[Blattmann et~al.(2023)Blattmann, Rombach, Ling, Dockhorn, Kim, Fidler,
  and Kreis]{blattmann2023align}
Andreas Blattmann, Robin Rombach, Huan Ling, Tim Dockhorn, Seung~Wook Kim,
  Sanja Fidler, and Karsten Kreis.
\newblock Align your latents: High-resolution video synthesis with latent
  diffusion models.
\newblock In \emph{CVPR}, 2023.

\bibitem[Brooks et~al.(2023)Brooks, Holynski, and
  Efros]{brooks2023instructpix2pix}
Tim Brooks, Aleksander Holynski, and Alexei~A Efros.
\newblock Instructpix2pix: Learning to follow image editing instructions.
\newblock In \emph{CVPR}, pages 18392--18402, 2023.

\bibitem[Chen et~al.(2023{\natexlab{a}})Chen, Zhang, Wang, Duan, Zhou, and
  Zhu]{chen2023disenbooth}
Hong Chen, Yipeng Zhang, Xin Wang, Xuguang Duan, Yuwei Zhou, and Wenwu Zhu.
\newblock Disenbooth: Identity-preserving disentangled tuning for
  subject-driven text-to-image generation.
\newblock \emph{arXiv preprint arXiv:2305.03374}, 2, 2023{\natexlab{a}}.

\bibitem[Chen et~al.(2023{\natexlab{b}})Chen, Hu, Li, Rui, Jia, Chang, and
  Cohen]{chen2023subject}
Wenhu Chen, Hexiang Hu, Yandong Li, Nataniel Rui, Xuhui Jia, Ming-Wei Chang,
  and William~W Cohen.
\newblock Subject-driven text-to-image generation via apprenticeship learning.
\newblock \emph{arXiv preprint arXiv:2304.00186}, 2023{\natexlab{b}}.

\bibitem[Cohen et~al.(2024)Cohen, Kulikov, Kleiner, Huberman-Spiegelglas, and
  Michaeli]{cohen2024slicedit}
Nathaniel Cohen, Vladimir Kulikov, Matan Kleiner, Inbar Huberman-Spiegelglas,
  and Tomer Michaeli.
\newblock Slicedit: Zero-shot video editing with text-to-image diffusion models
  using spatio-temporal slices.
\newblock In \emph{ICML}, pages 9109--9137. PMLR, 2024.

\bibitem[Cong et~al.(2023)Cong, Xu, Simon, Chen, Ren, Xie, Perez-Rua,
  Rosenhahn, Xiang, and He]{cong2023flatten}
Yuren Cong, Mengmeng Xu, Christian Simon, Shoufa Chen, Jiawei Ren, Yanping Xie,
  Juan-Manuel Perez-Rua, Bodo Rosenhahn, Tao Xiang, and Sen He.
\newblock Flatten: optical flow-guided attention for consistent text-to-video
  editing.
\newblock \emph{arXiv preprint arXiv:2310.05922}, 2023.

\bibitem[Dhariwal and Nichol(2021)]{dhariwal2021diffusion}
Prafulla Dhariwal and Alexander Nichol.
\newblock Diffusion models beat gans on image synthesis.
\newblock In \emph{NeurIPS}, pages 8780--8794, 2021.

\bibitem[Esser et~al.(2023)Esser, Chiu, Atighehchian, Granskog, and
  Germanidis]{esser2023structure}
Patrick Esser, Johnathan Chiu, Parmida Atighehchian, Jonathan Granskog, and
  Anastasis Germanidis.
\newblock Structure and content-guided video synthesis with diffusion models.
\newblock In \emph{ICCV}, pages 7346--7356, 2023.

\bibitem[Feng et~al.(2024)Feng, Weng, Wang, Yuan, Bao, Luo, Chen, and
  Guo]{feng2024ccedit}
Ruoyu Feng, Wenming Weng, Yanhui Wang, Yuhui Yuan, Jianmin Bao, Chong Luo,
  Zhibo Chen, and Baining Guo.
\newblock Ccedit: Creative and controllable video editing via diffusion models.
\newblock In \emph{CVPR}, pages 6712--6722, 2024.

\bibitem[Gal et~al.(2023)Gal, Alaluf, Atzmon, Patashnik, Bermano, Chechik, and
  Cohen-or]{gal2023image}
Rinon Gal, Yuval Alaluf, Yuval Atzmon, Or Patashnik, Amit~Haim Bermano, Gal
  Chechik, and Daniel Cohen-or.
\newblock An image is worth one word: Personalizing text-to-image generation
  using textual inversion.
\newblock In \emph{ICLR}, 2023.

\bibitem[Ge et~al.(2023)Ge, Nah, Liu, Poon, Tao, Catanzaro, Jacobs, Huang, Liu,
  and Balaji]{ge2023preserve}
Songwei Ge, Seungjun Nah, Guilin Liu, Tyler Poon, Andrew Tao, Bryan Catanzaro,
  David Jacobs, Jia-Bin Huang, Ming-Yu Liu, and Yogesh Balaji.
\newblock Preserve your own correlation: A noise prior for video diffusion
  models.
\newblock In \emph{ICCV}, 2023.

\bibitem[Geyer et~al.(2023)Geyer, Bar-Tal, Bagon, and
  Dekel]{geyer2023tokenflow}
Michal Geyer, Omer Bar-Tal, Shai Bagon, and Tali Dekel.
\newblock Tokenflow: Consistent diffusion features for consistent video
  editing.
\newblock \emph{arXiv preprint arXiv:2307.10373}, 2023.

\bibitem[Gu et~al.(2023)Gu, Wang, Zhao, Fu, Xiong, Liu, Zhang, Zhang, Zhang,
  Jung, et~al.]{gu2023photoswap}
Jing Gu, Yilin Wang, Nanxuan Zhao, Tsu-Jui Fu, Wei Xiong, Qing Liu, Zhifei
  Zhang, He Zhang, Jianming Zhang, HyunJoon Jung, et~al.
\newblock Photoswap: Personalized subject swapping in images.
\newblock \emph{arXiv preprint arXiv:2305.18286}, 2023.

\bibitem[Gu et~al.(2024{\natexlab{a}})Gu, Wang, Wu, Shi, Chen, Fan, Xiao, Zhao,
  Chang, Wu, et~al.]{gu2023mix}
Yuchao Gu, Xintao Wang, Jay~Zhangjie Wu, Yujun Shi, Yunpeng Chen, Zihan Fan,
  Wuyou Xiao, Rui Zhao, Shuning Chang, Weijia Wu, et~al.
\newblock Mix-of-show: Decentralized low-rank adaptation for multi-concept
  customization of diffusion models.
\newblock In \emph{NeurIPS}, 2024{\natexlab{a}}.

\bibitem[Gu et~al.(2024{\natexlab{b}})Gu, Zhou, Wu, Yu, Liu, Zhao, Wu, Zhang,
  Shou, and Tang]{gu2024videoswap}
Yuchao Gu, Yipin Zhou, Bichen Wu, Licheng Yu, Jia-Wei Liu, Rui Zhao,
  Jay~Zhangjie Wu, David~Junhao Zhang, Mike~Zheng Shou, and Kevin Tang.
\newblock Videoswap: Customized video subject swapping with interactive
  semantic point correspondence.
\newblock In \emph{CVPR}, pages 7621--7630, 2024{\natexlab{b}}.

\bibitem[Guo et~al.(2023)Guo, Yang, Rao, Liang, Wang, Qiao, Agrawala, Lin, and
  Dai]{guo2023animatediff}
Yuwei Guo, Ceyuan Yang, Anyi Rao, Zhengyang Liang, Yaohui Wang, Yu Qiao,
  Maneesh Agrawala, Dahua Lin, and Bo Dai.
\newblock Animatediff: Animate your personalized text-to-image diffusion models
  without specific tuning.
\newblock \emph{arXiv preprint arXiv:2307.04725}, 2023.

\bibitem[Hertz et~al.(2022)Hertz, Mokady, Tenenbaum, Aberman, Pritch, and
  Cohen-Or]{hertz2022prompt}
Amir Hertz, Ron Mokady, Jay Tenenbaum, Kfir Aberman, Yael Pritch, and Daniel
  Cohen-Or.
\newblock Prompt-to-prompt image editing with cross attention control.
\newblock \emph{arXiv preprint arXiv:2208.01626}, 2022.

\bibitem[Ho et~al.(2020)Ho, Jain, and Abbeel]{ho2020denoising}
Jonathan Ho, Ajay Jain, and Pieter Abbeel.
\newblock Denoising diffusion probabilistic models.
\newblock In \emph{NeurIPS}, pages 6840--6851, 2020.

\bibitem[Ho et~al.(2022{\natexlab{a}})Ho, Chan, Saharia, Whang, Gao, Gritsenko,
  Kingma, Poole, Norouzi, Fleet, et~al.]{ho2022imagen}
Jonathan Ho, William Chan, Chitwan Saharia, Jay Whang, Ruiqi Gao, Alexey
  Gritsenko, Diederik~P Kingma, Ben Poole, Mohammad Norouzi, David~J Fleet,
  et~al.
\newblock Imagen video: High definition video generation with diffusion models.
\newblock \emph{arXiv preprint arXiv:2210.02303}, 2022{\natexlab{a}}.

\bibitem[Ho et~al.(2022{\natexlab{b}})Ho, Saharia, Chan, Fleet, Norouzi, and
  Salimans]{ho2022cascaded}
Jonathan Ho, Chitwan Saharia, William Chan, David~J Fleet, Mohammad Norouzi,
  and Tim Salimans.
\newblock Cascaded diffusion models for high fidelity image generation.
\newblock \emph{JMLR}, 23\penalty0 (1):\penalty0 2249--2281,
  2022{\natexlab{b}}.

\bibitem[Ho et~al.(2022{\natexlab{c}})Ho, Salimans, Gritsenko, Chan, Norouzi,
  and Fleet]{ho2022video}
Jonathan Ho, Tim Salimans, Alexey Gritsenko, William Chan, Mohammad Norouzi,
  and David~J Fleet.
\newblock Video diffusion models.
\newblock In \emph{NeurIPS}, 2022{\natexlab{c}}.

\bibitem[Hu and Xu(2023)]{hu2023videocontrolnet}
Zhihao Hu and Dong Xu.
\newblock Videocontrolnet: A motion-guided video-to-video translation framework
  by using diffusion model with controlnet.
\newblock \emph{arXiv preprint arXiv:2307.14073}, 2023.

\bibitem[Huang et~al.(2023)Huang, Xie, Wang, Yuan, Cun, Ge, Zhou, Dong, Huang,
  Zhang, et~al.]{huang2023smartedit}
Yuzhou Huang, Liangbin Xie, Xintao Wang, Ziyang Yuan, Xiaodong Cun, Yixiao Ge,
  Jiantao Zhou, Chao Dong, Rui Huang, Ruimao Zhang, et~al.
\newblock Smartedit: Exploring complex instruction-based image editing with
  multimodal large language models.
\newblock \emph{arXiv preprint arXiv:2312.06739}, 2023.

\bibitem[Huang et~al.(2024{\natexlab{a}})Huang, Huang, Liu, Yan, Lv, Liu,
  Xiong, Zhang, Chen, and Cao]{huang2024diffusion}
Yi Huang, Jiancheng Huang, Yifan Liu, Mingfu Yan, Jiaxi Lv, Jianzhuang Liu, Wei
  Xiong, He Zhang, Shifeng Chen, and Liangliang Cao.
\newblock Diffusion model-based image editing: A survey.
\newblock \emph{arXiv preprint arXiv:2402.17525}, 2024{\natexlab{a}}.

\bibitem[Huang et~al.(2024{\natexlab{b}})Huang, He, Yu, Zhang, Si, Jiang,
  Zhang, Wu, Jin, Chanpaisit, et~al.]{huang2024vbench}
Ziqi Huang, Yinan He, Jiashuo Yu, Fan Zhang, Chenyang Si, Yuming Jiang, Yuanhan
  Zhang, Tianxing Wu, Qingyang Jin, Nattapol Chanpaisit, et~al.
\newblock Vbench: Comprehensive benchmark suite for video generative models.
\newblock In \emph{CVPR}, pages 21807--21818, 2024{\natexlab{b}}.

\bibitem[Kara et~al.(2024)Kara, Kurtkaya, Yesiltepe, Rehg, and
  Yanardag]{kara2024rave}
Ozgur Kara, Bariscan Kurtkaya, Hidir Yesiltepe, James~M Rehg, and Pinar
  Yanardag.
\newblock Rave: Randomized noise shuffling for fast and consistent video
  editing with diffusion models.
\newblock In \emph{CVPR}, pages 6507--6516, 2024.

\bibitem[Khachatryan et~al.(2023)Khachatryan, Movsisyan, Tadevosyan, Henschel,
  Wang, Navasardyan, and Shi]{khachatryan2023text2video}
Levon Khachatryan, Andranik Movsisyan, Vahram Tadevosyan, Roberto Henschel,
  Zhangyang Wang, Shant Navasardyan, and Humphrey Shi.
\newblock Text2video-zero: Text-to-image diffusion models are zero-shot video
  generators.
\newblock In \emph{ICCV}, 2023.

\bibitem[Kong et~al.(2024)Kong, Zhang, Yang, Wang, Zhang, Wu, Chen, Liu, and
  Luo]{kong2024omg}
Zhe Kong, Yong Zhang, Tianyu Yang, Tao Wang, Kaihao Zhang, Bizhu Wu, Guanying
  Chen, Wei Liu, and Wenhan Luo.
\newblock Omg: Occlusion-friendly personalized multi-concept generation in
  diffusion models.
\newblock \emph{arXiv preprint arXiv:2403.10983}, 2024.

\bibitem[Ku et~al.(2024)Ku, Wei, Ren, Yang, and Chen]{ku2024anyv2v}
Max Ku, Cong Wei, Weiming Ren, Huan Yang, and Wenhu Chen.
\newblock Anyv2v: A tuning-free framework for any video-to-video editing tasks.
\newblock \emph{TMLR}, 2024.

\bibitem[Kumari et~al.(2023)Kumari, Zhang, Zhang, Shechtman, and
  Zhu]{kumari2023multi}
Nupur Kumari, Bingliang Zhang, Richard Zhang, Eli Shechtman, and Jun-Yan Zhu.
\newblock Multi-concept customization of text-to-image diffusion.
\newblock In \emph{CVPR}, pages 1931--1941, 2023.

\bibitem[Li et~al.(2024{\natexlab{a}})Li, Zhang, Huang, Liu, Pei, Shao, and
  Xu]{li2024magiceraser}
Fan Li, Zixiao Zhang, Yi Huang, Jianzhuang Liu, Renjing Pei, Bin Shao, and
  Songcen Xu.
\newblock Magiceraser: Erasing any objects via semantics-aware control.
\newblock \emph{arXiv preprint arXiv:2410.10207}, 2024{\natexlab{a}}.

\bibitem[Li et~al.(2024{\natexlab{b}})Li, Li, Yang, Liu, Yue, Lin, and
  Xu]{li2024video}
Maomao Li, Yu Li, Tianyu Yang, Yunfei Liu, Dongxu Yue, Zhihui Lin, and Dong Xu.
\newblock A video is worth 256 bases: Spatial-temporal expectation-maximization
  inversion for zero-shot video editing.
\newblock In \emph{CVPR}, pages 7528--7537, 2024{\natexlab{b}}.

\bibitem[Li et~al.(2024{\natexlab{c}})Li, Ma, Yang, and Yang]{li2024vidtome}
Xirui Li, Chao Ma, Xiaokang Yang, and Ming-Hsuan Yang.
\newblock Vidtome: Video token merging for zero-shot video editing.
\newblock In \emph{CVPR}, 2024{\natexlab{c}}.

\bibitem[Liang et~al.(2024)Liang, Wu, Wang, Yu, Li, Zhao, Misra, Huang, Zhang,
  Vajda, et~al.]{liang2024flowvid}
Feng Liang, Bichen Wu, Jialiang Wang, Licheng Yu, Kunpeng Li, Yinan Zhao, Ishan
  Misra, Jia-Bin Huang, Peizhao Zhang, Peter Vajda, et~al.
\newblock Flowvid: Taming imperfect optical flows for consistent video-to-video
  synthesis.
\newblock In \emph{CVPR}, pages 8207--8216, 2024.

\bibitem[Liew et~al.(2023)Liew, Yan, Zhang, Xu, and Feng]{liew2023magicedit}
Jun~Hao Liew, Hanshu Yan, Jianfeng Zhang, Zhongcong Xu, and Jiashi Feng.
\newblock Magicedit: High-fidelity and temporally coherent video editing.
\newblock \emph{arXiv preprint arXiv:2308.14749}, 2023.

\bibitem[Liu et~al.(2024)Liu, Zhang, Li, Lin, and Jia]{liu2024video}
Shaoteng Liu, Yuechen Zhang, Wenbo Li, Zhe Lin, and Jiaya Jia.
\newblock Video-p2p: Video editing with cross-attention control.
\newblock In \emph{CVPR}, pages 8599--8608, 2024.

\bibitem[Liu et~al.(2023)Liu, Feng, Zhu, Zhang, Zheng, Liu, Zhao, Zhou, and
  Cao]{liu2023cones}
Zhiheng Liu, Ruili Feng, Kai Zhu, Yifei Zhang, Kecheng Zheng, Yu Liu, Deli
  Zhao, Jingren Zhou, and Yang Cao.
\newblock Cones: Concept neurons in diffusion models for customized generation.
\newblock In \emph{ICML}, 2023.

\bibitem[Lu et~al.(2023)Lu, Tunanyan, Wang, Navasardyan, Wang, and
  Shi]{lu2023specialist}
Haoming Lu, Hazarapet Tunanyan, Kai Wang, Shant Navasardyan, Zhangyang Wang,
  and Humphrey Shi.
\newblock Specialist diffusion: Plug-and-play sample-efficient fine-tuning of
  text-to-image diffusion models to learn any unseen style.
\newblock In \emph{CVPR}, pages 14267--14276, 2023.

\bibitem[Lv et~al.(2024)Lv, Huang, Yan, Huang, Liu, Liu, Wen, Chen, and
  Chen]{lv2024gpt4motion}
Jiaxi Lv, Yi Huang, Mingfu Yan, Jiancheng Huang, Jianzhuang Liu, Yifan Liu,
  Yafei Wen, Xiaoxin Chen, and Shifeng Chen.
\newblock Gpt4motion: Scripting physical motions in text-to-video generation
  via blender-oriented gpt planning.
\newblock In \emph{CVPRW}, pages 1430--1440, 2024.

\bibitem[Melas-Kyriazi et~al.(2022)Melas-Kyriazi, Rupprecht, Laina, and
  Vedaldi]{melas2022deep}
Luke Melas-Kyriazi, Christian Rupprecht, Iro Laina, and Andrea Vedaldi.
\newblock Deep spectral methods: A surprisingly strong baseline for
  unsupervised semantic segmentation and localization.
\newblock In \emph{CVPR}, pages 8364--8375, 2022.

\bibitem[Mou et~al.(2023)Mou, Wang, Xie, Zhang, Qi, Shan, and Qie]{mou2023t2i}
Chong Mou, Xintao Wang, Liangbin Xie, Jian Zhang, Zhongang Qi, Ying Shan, and
  Xiaohu Qie.
\newblock T2i-adapter: Learning adapters to dig out more controllable ability
  for text-to-image diffusion models.
\newblock \emph{arXiv preprint arXiv:2302.08453}, 2023.

\bibitem[Oquab et~al.(2023)Oquab, Darcet, Moutakanni, Vo, Szafraniec, Khalidov,
  Fernandez, Haziza, Massa, El-Nouby, et~al.]{oquab2023dinov2}
Maxime Oquab, Timoth{\'e}e Darcet, Th{\'e}o Moutakanni, Huy Vo, Marc
  Szafraniec, Vasil Khalidov, Pierre Fernandez, Daniel Haziza, Francisco Massa,
  Alaaeldin El-Nouby, et~al.
\newblock Dinov2: Learning robust visual features without supervision.
\newblock \emph{arXiv preprint arXiv:2304.07193}, 2023.

\bibitem[Ouyang et~al.(2024)Ouyang, Dong, Yang, Si, and Pan]{ouyang2024i2vedit}
Wenqi Ouyang, Yi Dong, Lei Yang, Jianlou Si, and Xingang Pan.
\newblock I2vedit: First-frame-guided video editing via image-to-video
  diffusion models.
\newblock In \emph{SIGGRAPH Asia 2024 Conference Papers}, pages 1--11, 2024.

\bibitem[Perazzi et~al.(2016)Perazzi, Pont-Tuset, McWilliams, Van~Gool, Gross,
  and Sorkine-Hornung]{perazzi2016benchmark}
Federico Perazzi, Jordi Pont-Tuset, Brian McWilliams, Luc Van~Gool, Markus
  Gross, and Alexander Sorkine-Hornung.
\newblock A benchmark dataset and evaluation methodology for video object
  segmentation.
\newblock In \emph{CVPR}, pages 724--732, 2016.

\bibitem[Qi et~al.(2023)Qi, Cun, Zhang, Lei, Wang, Shan, and
  Chen]{qi2023fatezero}
Chenyang Qi, Xiaodong Cun, Yong Zhang, Chenyang Lei, Xintao Wang, Ying Shan,
  and Qifeng Chen.
\newblock Fatezero: Fusing attentions for zero-shot text-based video editing.
\newblock In \emph{ICCV}, pages 15932--15942, 2023.

\bibitem[Rombach et~al.(2022)Rombach, Blattmann, Lorenz, Esser, and
  Ommer]{rombach2022high}
Robin Rombach, Andreas Blattmann, Dominik Lorenz, Patrick Esser, and Bj{\"o}rn
  Ommer.
\newblock High-resolution image synthesis with latent diffusion models.
\newblock In \emph{CVPR}, 2022.

\bibitem[Ruiz et~al.(2023)Ruiz, Li, Jampani, Pritch, Rubinstein, and
  Aberman]{ruiz2023dreambooth}
Nataniel Ruiz, Yuanzhen Li, Varun Jampani, Yael Pritch, Michael Rubinstein, and
  Kfir Aberman.
\newblock Dreambooth: Fine tuning text-to-image diffusion models for
  subject-driven generation.
\newblock In \emph{CVPR}, pages 22500--22510, 2023.

\bibitem[Ruiz et~al.(2024)Ruiz, Li, Jampani, Wei, Hou, Pritch, Wadhwa,
  Rubinstein, and Aberman]{ruiz2024hyperdreambooth}
Nataniel Ruiz, Yuanzhen Li, Varun Jampani, Wei Wei, Tingbo Hou, Yael Pritch,
  Neal Wadhwa, Michael Rubinstein, and Kfir Aberman.
\newblock Hyperdreambooth: Hypernetworks for fast personalization of
  text-to-image models.
\newblock In \emph{CVPR}, pages 6527--6536, 2024.

\bibitem[Saharia et~al.(2022)Saharia, Chan, Saxena, Li, Whang, Denton,
  Ghasemipour, Ayan, Mahdavi, Lopes, et~al.]{saharia2022photorealistic}
Chitwan Saharia, William Chan, Saurabh Saxena, Lala Li, Jay Whang, Emily
  Denton, Seyed Kamyar~Seyed Ghasemipour, Burcu~Karagol Ayan, S~Sara Mahdavi,
  Rapha~Gontijo Lopes, et~al.
\newblock Photorealistic text-to-image diffusion models with deep language
  understanding.
\newblock In \emph{NeurIPS}, 2022.

\bibitem[Sheynin et~al.(2023)Sheynin, Polyak, Singer, Kirstain, Zohar, Ashual,
  Parikh, and Taigman]{sheynin2023emu}
Shelly Sheynin, Adam Polyak, Uriel Singer, Yuval Kirstain, Amit Zohar, Oron
  Ashual, Devi Parikh, and Yaniv Taigman.
\newblock Emu edit: Precise image editing via recognition and generation tasks.
\newblock \emph{arXiv preprint arXiv:2311.10089}, 2023.

\bibitem[Shi et~al.(2023)Shi, Xiong, Lin, and Jung]{shi2023instantbooth}
Jing Shi, Wei Xiong, Zhe Lin, and Hyun~Joon Jung.
\newblock Instantbooth: Personalized text-to-image generation without test-time
  finetuning.
\newblock \emph{arXiv preprint arXiv:2304.03411}, 2023.

\bibitem[Shtedritski et~al.(2023)Shtedritski, Vedaldi, and
  Rupprecht]{shtedritski2023learning}
Aleksandar Shtedritski, Andrea Vedaldi, and Christian Rupprecht.
\newblock Learning universal semantic correspondences with no supervision and
  automatic data curation.
\newblock In \emph{ICCVW}, pages 933--943, 2023.

\bibitem[Singer et~al.(2023)Singer, Polyak, Hayes, Yin, An, Zhang, Hu, Yang,
  Ashual, Gafni, et~al.]{singer2023make}
Uriel Singer, Adam Polyak, Thomas Hayes, Xi Yin, Jie An, Songyang Zhang, Qiyuan
  Hu, Harry Yang, Oron Ashual, Oran Gafni, et~al.
\newblock Make-a-video: Text-to-video generation without text-video data.
\newblock In \emph{ICLR}, 2023.

\bibitem[Sohl-Dickstein et~al.(2015)Sohl-Dickstein, Weiss, Maheswaranathan, and
  Ganguli]{sohl2015deep}
Jascha Sohl-Dickstein, Eric Weiss, Niru Maheswaranathan, and Surya Ganguli.
\newblock Deep unsupervised learning using nonequilibrium thermodynamics.
\newblock In \emph{ICML}, 2015.

\bibitem[Song et~al.(2021{\natexlab{a}})Song, Meng, and
  Ermon]{song2021denoising}
Jiaming Song, Chenlin Meng, and Stefano Ermon.
\newblock Denoising diffusion implicit models.
\newblock In \emph{ICLR}, 2021{\natexlab{a}}.

\bibitem[Song and Ermon(2019)]{song2019generative}
Yang Song and Stefano Ermon.
\newblock Generative modeling by estimating gradients of the data distribution.
\newblock In \emph{NeurIPS}, 2019.

\bibitem[Song et~al.(2021{\natexlab{b}})Song, Sohl-Dickstein, Kingma, Kumar,
  Ermon, and Poole]{song2021score}
Yang Song, Jascha Sohl-Dickstein, Diederik~P Kingma, Abhishek Kumar, Stefano
  Ermon, and Ben Poole.
\newblock Score-based generative modeling through stochastic differential
  equations.
\newblock In \emph{ICLR}, 2021{\natexlab{b}}.

\bibitem[Sun et~al.(2024)Sun, Tu, Liao, and Tao]{sun2024diffusion}
Wenhao Sun, Rong-Cheng Tu, Jingyi Liao, and Dacheng Tao.
\newblock Diffusion model-based video editing: A survey.
\newblock \emph{arXiv preprint arXiv:2407.07111}, 2024.

\bibitem[Tang et~al.(2023)Tang, Jia, Wang, Phoo, and
  Hariharan]{tang2023emergent}
Luming Tang, Menglin Jia, Qianqian Wang, Cheng~Perng Phoo, and Bharath
  Hariharan.
\newblock Emergent correspondence from image diffusion.
\newblock In \emph{NeurIPS}, pages 1363--1389, 2023.

\bibitem[Tewel et~al.(2023)Tewel, Gal, Chechik, and Atzmon]{tewel2023key}
Yoad Tewel, Rinon Gal, Gal Chechik, and Yuval Atzmon.
\newblock Key-locked rank one editing for text-to-image personalization.
\newblock In \emph{ACM SIGGRAPH}, pages 1--11, 2023.

\bibitem[Voronov et~al.(2023)Voronov, Khoroshikh, Babenko, and
  Ryabinin]{voronov2023loss}
Anton Voronov, Mikhail Khoroshikh, Artem Babenko, and Max Ryabinin.
\newblock Is this loss informative? faster text-to-image customization by
  tracking objective dynamics.
\newblock In \emph{NeurIPS}, 2023.

\bibitem[Wang et~al.(2024)Wang, Ma, Guo, Xiao, Huang, and Li]{wang2024cove}
Jiangshan Wang, Yue Ma, Jiayi Guo, Yicheng Xiao, Gao Huang, and Xiu Li.
\newblock Cove: Unleashing the diffusion feature correspondence for consistent
  video editing.
\newblock \emph{arXiv preprint arXiv:2406.08850}, 2024.

\bibitem[Wang et~al.(2023{\natexlab{a}})Wang, Chen, Ma, Zhou, Huang, Wang,
  Yang, He, Yu, Yang, et~al.]{wang2023lavie}
Yaohui Wang, Xinyuan Chen, Xin Ma, Shangchen Zhou, Ziqi Huang, Yi Wang, Ceyuan
  Yang, Yinan He, Jiashuo Yu, Peiqing Yang, et~al.
\newblock Lavie: High-quality video generation with cascaded latent diffusion
  models.
\newblock \emph{arXiv preprint arXiv:2309.15103}, 2023{\natexlab{a}}.

\bibitem[Wang et~al.(2023{\natexlab{b}})Wang, He, Li, Li, Yu, Ma, Chen, Wang,
  Luo, Liu, et~al.]{wang2023internvid}
Yi Wang, Yinan He, Yizhuo Li, Kunchang Li, Jiashuo Yu, Xin Ma, Xinyuan Chen,
  Yaohui Wang, Ping Luo, Ziwei Liu, et~al.
\newblock Internvid: A large-scale video-text dataset for multimodal
  understanding and generation.
\newblock \emph{arXiv preprint arXiv:2307.06942}, 2023{\natexlab{b}}.

\bibitem[Wei et~al.(2023)Wei, Zhang, Ji, Bai, Zhang, and Zuo]{wei2023elite}
Yuxiang Wei, Yabo Zhang, Zhilong Ji, Jinfeng Bai, Lei Zhang, and Wangmeng Zuo.
\newblock Elite: Encoding visual concepts into textual embeddings for
  customized text-to-image generation.
\newblock \emph{arXiv preprint arXiv:2302.13848}, 2023.

\bibitem[Wei et~al.(2024)Wei, Zhang, Qing, Yuan, Liu, Liu, Zhang, Zhou, and
  Shan]{wei2024dreamvideo}
Yujie Wei, Shiwei Zhang, Zhiwu Qing, Hangjie Yuan, Zhiheng Liu, Yu Liu, Yingya
  Zhang, Jingren Zhou, and Hongming Shan.
\newblock Dreamvideo: Composing your dream videos with customized subject and
  motion.
\newblock In \emph{CVPR}, pages 6537--6549, 2024.

\bibitem[Wu et~al.(2023{\natexlab{a}})Wu, Ge, Wang, Lei, Gu, Shi, Hsu, Shan,
  Qie, and Shou]{wu2023tune}
Jay~Zhangjie Wu, Yixiao Ge, Xintao Wang, Stan~Weixian Lei, Yuchao Gu, Yufei
  Shi, Wynne Hsu, Ying Shan, Xiaohu Qie, and Mike~Zheng Shou.
\newblock Tune-a-video: One-shot tuning of image diffusion models for
  text-to-video generation.
\newblock In \emph{CVPR}, pages 7623--7633, 2023{\natexlab{a}}.

\bibitem[Wu et~al.(2023{\natexlab{b}})Wu, Li, Gao, Dong, Bai, Singh, Xiang, Li,
  Huang, Sun, et~al.]{wu2023cvpr}
Jay~Zhangjie Wu, Xiuyu Li, Difei Gao, Zhen Dong, Jinbin Bai, Aishani Singh,
  Xiaoyu Xiang, Youzeng Li, Zuwei Huang, Yuanxi Sun, et~al.
\newblock Cvpr 2023 text guided video editing competition.
\newblock \emph{arXiv preprint arXiv:2310.16003}, 2023{\natexlab{b}}.

\bibitem[Xiao et~al.(2023)Xiao, Yin, Gong, Zang, Ren, and Yuan]{xiao2023comcat}
Jinqi Xiao, Miao Yin, Yu Gong, Xiao Zang, Jian Ren, and Bo Yuan.
\newblock Comcat: towards efficient compression and customization of
  attention-based vision models.
\newblock \emph{arXiv preprint arXiv:2305.17235}, 2023.

\bibitem[Xing et~al.(2023)Xing, Feng, Chen, Dai, Hu, Xu, Wu, and
  Jiang]{xing2023survey}
Zhen Xing, Qijun Feng, Haoran Chen, Qi Dai, Han Hu, Hang Xu, Zuxuan Wu, and
  Yu-Gang Jiang.
\newblock A survey on video diffusion models.
\newblock \emph{arXiv preprint arXiv:2310.10647}, 2023.

\bibitem[Xue et~al.(2022)Xue, Hang, Zeng, Sun, Liu, Yang, Fu, and
  Guo]{xue2022advancing}
Hongwei Xue, Tiankai Hang, Yanhong Zeng, Yuchong Sun, Bei Liu, Huan Yang,
  Jianlong Fu, and Baining Guo.
\newblock Advancing high-resolution video-language representation with
  large-scale video transcriptions.
\newblock In \emph{CVPR}, 2022.

\bibitem[Yang et~al.(2023)Yang, Zhou, Liu, and Loy]{yang2023rerender}
Shuai Yang, Yifan Zhou, Ziwei Liu, and Chen~Change Loy.
\newblock Rerender a video: Zero-shot text-guided video-to-video translation.
\newblock In \emph{SIGGRAPH Asia 2023 Conference Papers}, pages 1--11, 2023.

\bibitem[Ye et~al.(2023)Ye, Zhang, Liu, Han, and Yang]{ye2023ip}
Hu Ye, Jun Zhang, Sibo Liu, Xiao Han, and Wei Yang.
\newblock Ip-adapter: Text compatible image prompt adapter for text-to-image
  diffusion models.
\newblock \emph{arXiv preprint arXiv:2308.06721}, 2023.

\bibitem[Yuan et~al.(2023)Yuan, Cao, Wang, Qi, Yuan, and
  Shan]{yuan2023customnet}
Ziyang Yuan, Mingdeng Cao, Xintao Wang, Zhongang Qi, Chun Yuan, and Ying Shan.
\newblock Customnet: Zero-shot object customization with variable-viewpoints in
  text-to-image diffusion models.
\newblock \emph{arXiv preprint arXiv:2310.19784}, 2023.

\bibitem[Zhang et~al.(2023{\natexlab{a}})Zhang, Herrmann, Hur, Polania~Cabrera,
  Jampani, Sun, and Yang]{zhang2023tale}
Junyi Zhang, Charles Herrmann, Junhwa Hur, Luisa Polania~Cabrera, Varun
  Jampani, Deqing Sun, and Ming-Hsuan Yang.
\newblock A tale of two features: Stable diffusion complements dino for
  zero-shot semantic correspondence.
\newblock In \emph{NeurIPS}, 2023{\natexlab{a}}.

\bibitem[Zhang et~al.(2023{\natexlab{b}})Zhang, Rao, and
  Agrawala]{zhang2023adding}
Lvmin Zhang, Anyi Rao, and Maneesh Agrawala.
\newblock Adding conditional control to text-to-image diffusion models.
\newblock In \emph{ICCV}, pages 3836--3847, 2023{\natexlab{b}}.

\bibitem[Zhang et~al.(2024)Zhang, Li, Nie, Han, Guo, and Liu]{zhang2024towards}
Zicheng Zhang, Bonan Li, Xuecheng Nie, Congying Han, Tiande Guo, and Luoqi Liu.
\newblock Towards consistent video editing with text-to-image diffusion models.
\newblock In \emph{NeurIPS}, 2024.

\bibitem[Zhao et~al.(2023)Zhao, Wang, Bao, Li, and Zhu]{zhao2023controlvideo}
Min Zhao, Rongzhen Wang, Fan Bao, Chongxuan Li, and Jun Zhu.
\newblock Controlvideo: Adding conditional control for one shot text-to-video
  editing.
\newblock \emph{arXiv preprint arXiv:2305.17098}, 2\penalty0 (3), 2023.

\end{thebibliography}
}


\end{document}